\documentclass[conference]{IEEEtran}
\IEEEoverridecommandlockouts
\usepackage{cite}
\usepackage{amsmath,amssymb,amsfonts}
\usepackage{algorithmic}
\usepackage{graphicx}
\usepackage{textcomp}
\usepackage{xcolor}
\usepackage{booktabs}
\usepackage{multirow}
\usepackage{url}

\newcommand{\Sz}{S_{0}}                   
\newcommand{\So}{S_{1}}                   
\newcommand{\Ggap}{G}                     
\newcommand{\Aset}{\mathcal{A}}           
\newcommand{\Vset}{\mathcal{V}}           
\newcommand{\Wset}{\mathcal{W}}           
\newcommand{\Pver}{\pi(\Vset)}            
\newcommand{\Pact}{\pi(\Wset)}            
\newcommand{\Presp}{\pi_{r}}              
\newcommand{\otrue}{o}                    
\newcommand{\oinj}{\tilde{o}}             
\newcommand{\amax}{a^{\star}}             
\newcommand{\koff}{k}                     

\newcommand{\Acl}{\ensuremath{\mathsf{C}}}                    
\newcommand{\Afail}{\ensuremath{\mathsf{F}}}                  
\newcommand{\Astat}{\ensuremath{\mathsf{N}_{\mathsf{s}}}}     
\newcommand{\Apay}{\ensuremath{\mathsf{N}_{\mathsf{m}}}}      

\begin{document}

\title{Robustness Analysis of Agentic AI to Inconsistent\\
and Incomplete Tool Responses}

\author{\IEEEauthorblockN{Jiachen Xu}
\IEEEauthorblockA{\textit{Department of Computer Science} \\
\textit{Aalborg University}\\
Aalborg, Denmark \\
jiachenxu@cs.aau.dk}
\and
\IEEEauthorblockN{Torben Bach Pedersen}
\IEEEauthorblockA{\textit{Department of Computer Science} \\
\textit{Aalborg University}\\
Aalborg, Denmark \\
tbp@cs.aau.dk}
\and
\IEEEauthorblockN{Zhongming Yao}
\IEEEauthorblockA{\textit{College of Computer Science and Technology} \\
\textit{Zhejiang University}\\
Hangzhou, China \\
yaozzzm@gmail.com }
\and
\IEEEauthorblockN{Xiaoyu Zhang}
\IEEEauthorblockA{\textit{School of Information Science and Engineering} \\
\textit{Northeastern University}\\
Shenyang, China \\
2410400@stu.neu.edu.cn}
\and
\IEEEauthorblockN{Yushuai Li}
\IEEEauthorblockA{\textit{Department of Computer Science} \\
\textit{Aalborg University}\\
Aalborg, Denmark \\
yusli@cs.aau.dk}
}

\maketitle

\pagestyle{plain}
\thispagestyle{plain}

\begin{abstract}
Tool-using agents increasingly rely on external tools to complete multi-step tasks, but tool returns can fail in different ways and require different recovery actions. Existing robustness studies often use uncertainty-based measures to detect when an agent becomes unreliable. These measures can reveal that something has gone wrong, but they do not directly identify the type of tool failure or the appropriate response. We address this limitation by analyzing tool failures at the moment a return enters the agent context. Our approach combines two complementary signals. The first compares the likelihood of the returned content under the tool schema and under the full trajectory prefix. The second measures the agent's probability distribution over its legal next actions. We evaluate the approach by injecting incomplete and inconsistent returns into a retail customer-service benchmark. The results show that likelihood-based signals clearly capture incomplete returns and some direct inconsistencies, while action-based signals reveal how strongly a failure changes the next decision. Some failures that are weak under likelihood signals can still redirect the agent toward state-changing actions. These findings show that tool failures can be recognized at the return boundary, but reliable diagnosis requires combining multiple signals.
\end{abstract}


\section{Introduction}

A robust tool-using agent does not need to avoid every bad tool return. Instead, it should respond to each bad return in a suitable way. The right response depends on how the tool fails. Robustness therefore starts with a basic question. When a tool return arrives, can the agent identify what kind of return it has received? Tool-using language agents are brittle under realistic noise. The diagnostic reported for this brittleness is step-wise reasoning entropy~\cite{agentnoisebench,actundernoise}: (i) on clean trajectories~\cite{L2}, entropy decreases as the agent converges on a solution, while (ii) on trajectories affected by tool noise~\cite{L3}, entropy increases from the middle of the episode onward. These results also show that agents with explicit reasoning can be \emph{less} robust to tool noise. They may accept corrupted evidence and continue building a fluent reasoning chain from it. A rising entropy curve can therefore indicate that something has gone wrong, but it does not explain what failure has occurred.

Noise, however, can take different forms~\cite{agentnoisebench,actundernoise}. A response is \emph{incomplete} when some evidence is missing because the call fails or the returned payload is truncated. In this case, the problem is often visible from the return itself, such as a missing field or an error message. A response is \emph{inconsistent} when it is well formed and internally coherent but conflicts with external information, such as the user's stated intent or an earlier tool return~\cite{toolhalluc,toolbehonest}. Such a return may appear valid on its own, even though its content is incorrect in the current context. These two failures require different responses. Incompleteness often calls for a retry or graceful degradation, while inconsistency calls for checking the claim against an independent source.

Existing methods usually rely on uncertainty-based measures to identify unreliable generations. Predictive entropy for structured prediction~\cite{predentropy} measures uncertainty over possible outputs, while semantic entropy~\cite{semanticentropy} groups meaning-equivalent samples before measuring dispersion. These methods can identify unreliable generations from the model's own distribution. However, they mainly indicate whether uncertainty is present rather than which type of tool failure caused it. A single curve~\cite{L5,L7} that summarizes a completed episode also cannot tell the agent which recovery action to take, especially when the curve remains flat. This limitation matters because different failures require different responses. Applying the wrong response can be as harmful as taking no action.


To address this problem, we examine tool returns at the moment they enter a ReAct-style loop~\cite{react}. Our goal is to identify the failure type before the agent continues reasoning. The key idea is that incomplete and inconsistent returns are abnormal with respect to different references. An incomplete return may be unlikely under the tool schema because it does not match the expected form of a valid return. In contrast, an inconsistent return may be fully schema-valid but unlikely given the preceding trajectory~\cite{L6}, including the user's intent and earlier tool returns. We therefore score each return from two views, once under the tool schema and once under the full interaction prefix. We also examine the agent's distribution over legal actions at the same boundary. In addition to entropy, we consider the peak probability and the tool group that receives the largest probability mass.

We evaluate these signals on the retail domain of $\tau^2$-bench~\cite{tau2bench}. We construct incomplete and schema-valid inconsistent returns while keeping the preceding trajectory unchanged. For incomplete failures, we use error strings produced by the environment. For inconsistent failures, we modify one field while keeping the payload schema-valid. We further consider two forms of inconsistency. In one case, the correct value already appears explicitly in the prefix. In the other, the contradiction can only be identified through the domain policy.

The results show that the two views capture different aspects of failure. The likelihood comparison clearly responds to incomplete returns and to inconsistencies that directly conflict with values already present in the context. However, it does not reliably capture inconsistencies that require policy reasoning. The action distribution responds to all three conditions, but its change reflects how strongly the faulty return affects the next action rather than the failure type itself. In particular, some inconsistencies that are weak under the likelihood signal can still push the agent toward a state-changing tool call instead of another read operation. These results show that the failure signal is already present at the tool-return boundary, but no single quantity captures all of its aspects.

\section{Measurement Design}
\label{sec:setup}

\subsection{Setting and notation}
\label{sec:setting}
We use the retail domain of $\tau^2$-bench~\cite{tau2bench,taubench}, which contains 114 tasks and 16 tools. The agent follows a ReAct-style loop~\cite{react}. At each \emph{decision point}, it either calls one of the 16 tools or replies to the user. After each tool call, the returned result is added to the context before the next decision point.

Let $i$ denote the decision point immediately before the injection site. Let $\tau_{<i}$ denote the trajectory prefix before this point, including the domain policy, user turns, and all earlier tool calls and returns. We use $u$ to denote the user's stated intent in this prefix. Let $\otrue$ be the original return of the tool call at step $i$, and let $\oinj$ be the return inserted in its place. We use $|\cdot|$ to denote the number of tokens in a string.
We evaluate each returned string under two conditioning contexts:
\begin{equation}
c_0 = \{\text{tool schema}\},
\qquad
c_1 = c_0 \cup \{\tau_{<i},\, u\},
\label{eq:contexts}
\end{equation}
where $c_0$ contains only the tool schema, and $c_1$ additionally includes the preceding trajectory and the user's intent. By construction, $c_0$ is a subset of $c_1$.

Every quantity below is computed from log-probabilities under a fixed prefix. After an injection, we obtain the action distribution by adding the injected return to the context and scoring all candidate actions at the next decision point. We measure only this single decision point, so $\pi$ is computed directly rather than estimated from empirical frequencies. We generate prefixes by replaying the reference action sequence of each task against the real domain database. This gives all settings the same prefix at decision point $i$. Each injected case is then compared with a clean case that differs in only one tool return.

\subsection{Quantities}
\label{sec:signals}

\emph{Two conditional surprisals.} A returned string $x$ is scored under each of
the two conditionings of Eq.~\eqref{eq:contexts}, normalized per token, and the
two are compared:
\begin{align}
S_k(x) &= -\tfrac{1}{|x|}\log P(x \mid c_k), \qquad k \in \{0,1\},
   \label{eq:surprisal}\\
\Ggap(x) &= \So(x) - \Sz(x) , \label{eq:gap}
\end{align}
where $\Sz$ reads the string against the tool schema alone, $\So$ reads it
against the whole prefix, and $\Ggap$ is therefore what the trajectory adds to
the schema.

\emph{Action-distribution entropy.} The action set $\Aset$ contains $K=17$ actions, including 16 tool calls and one reply action. We obtain the probability of each action directly from the model. The next-token distribution gives the probability of starting a tool call, and the following tool-name probabilities divide this mass across the 16 tools. The remaining probability is assigned to replying to the user. This gives a probability vector $\pi$ over $\Aset$, from which we compute the entropy.
\begin{equation}
H(\pi) = -\sum_{j=1}^{K} \pi_j \log \pi_j ,
\label{eq:entropy}
\end{equation}
where $\pi_j$ is the probability of the $j$-th action. We define entropy over the legal action set rather than vocabulary tokens. This avoids several problems with token-level entropy, including high uncertainty in free-form reasoning text, very low uncertainty in fixed formatting, and arguments copied directly from the context. The resulting entropy is bounded by $\log K$ and corresponds to the policy entropy used in reinforcement learning~\cite{agenticrl}. Entropy captures the shape of $\pi$, but not which action has the highest probability. We therefore record this information separately. Let $\amax = \arg\max_j \pi_j$ denote the most likely action, and let $a$ denote the reference action at that step. We set $A=1$ when $\amax=a$, and $A=0$ otherwise.

\emph{Grouped action probabilities.} We also record how the probability mass is distributed across different action groups. Let $\Pver$ denote the total probability assigned to the two tools that re-read mutable state, $\Pact$ the probability assigned to the seven write actions defined by the benchmark, and $\Presp$ the probability assigned to replying to the user. A decrease in $\Pver$ therefore means that the agent is less likely to verify the current state again.

\emph{Paired contrast.} Entropy drifts as a conversation lengthens and the
surprisals are model-subjective, so every quantity is reported relative to the
paired clean row of the same task at the same step. For any quantity $X$,
\begin{equation}
\Delta X = X(\oinj) - X(\otrue) ,
\label{eq:paired}
\end{equation}
the injected value minus the true one. Applied to Eq.~\eqref{eq:entropy} this is
$\Delta H = H(\pi \mid \oinj) - H(\pi \mid \otrue)$, in which the pre-injection
term cancels exactly. Applied to $A$ it takes the values $-1$, $0$ and $+1$,
recording a leading action displaced from the reference one, left where it was,
or moved onto it.

\emph{How the signs read.} A tampered field can still be syntactically valid and type-correct. It may therefore appear normal under $\Sz$, while its anomaly becomes visible only under $\So$, where the trajectory includes the user's statements and earlier tool returns. An execution failure shows the opposite pattern because it can already be unlikely under the tool schema. Since additional context usually makes a return easier to predict, $\Ggap$ should be clearly negative for clean returns. A contradiction should move it closer to zero. A negative $\Delta H$ means that the injected return makes the action distribution more concentrated, while a positive value makes it more dispersed. However, concentration alone does not reveal whether the agent is behaving correctly. It can reflect either false convergence or a confident retry. We therefore interpret $\Delta H$ together with the most likely action and the distribution of probability mass across action groups.

\subsection{Injection design}
\label{sec:protocol}
The injection site is the first call to \texttt{get\_order\_details} at position
two or later, this return being the most information-dense payload in the domain
and the site being required to fall mid-trajectory. Fixing the site by
\emph{tool} rather than by index holds the field vocabulary constant across
tasks; the index varies, but it does not act as a confound, because each row is
differenced against its own clean prefix at the same index.

Four arms are constructed at that site: a clean arm \Acl, a failed-return arm
\Afail{} standing for the incomplete family, and two inconsistent arms \Astat{}
and \Apay, whose shared letter marks the family and whose subscript names the
field that is falsified. These labels are used in place of the full names
throughout. \Acl{} leaves $\otrue$ in place. \Afail{} replaces the payload with
the error string the environment itself produces, the limiting case of an
incomplete response in which nothing of the payload survives. The inconsistent
family is
realized twice, on two different fields of the same payload: \Astat{} rewrites
the order status to another legal value from its seven-value enumeration, which
changes which actions the domain policy permits; \Apay{} rewrites a
payment-method identifier to one that is well-formed but does not belong to this
user. The two differ in how the contradiction is available: in \Apay{} the true value appears verbatim earlier in the prefix, so the conflict is one of tokens, whereas in \Astat{} the user has described the state of the order in prose and the conflict has to be reached through meaning.

\begin{figure*}[!t]
\centering
\includegraphics[width=\textwidth]{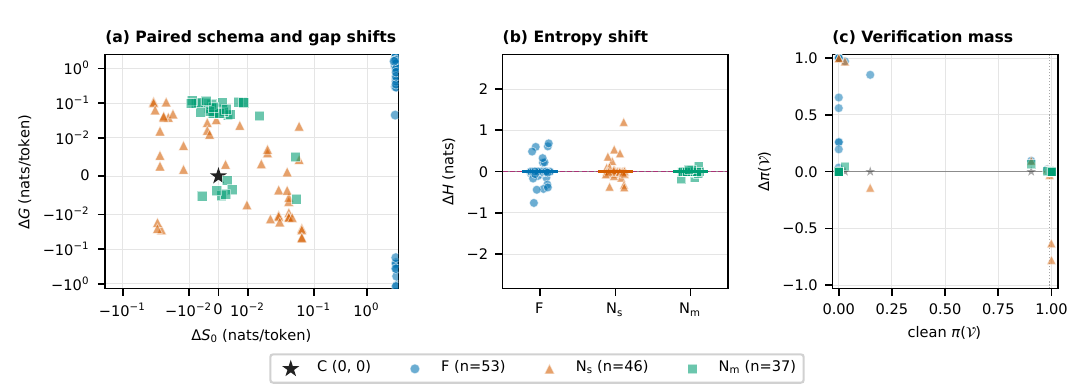}
\caption{Readings at the injection site. \Afail{} is the failed return, \Astat{} and \Apay{} the falsified order status and payment identifier (Section~\ref{sec:protocol}). Every quantity is paired against the same task's \Acl{} row, which therefore sits at the origin of (a) and on $y=0$ in (c).
\textbf{(a)} The $(\Delta\Sz,\Delta\Ggap)$ plane, both axes symmetric-log.
\textbf{(b)} $\Delta H$ per arm against the full $\pm\log 17$ range of the action
entropy, medians as horizontal bars; the dashed line, at minus the median clean
entropy, marks the furthest a median row could fall. \textbf{(c)} Paired change
in $\Pver$ against its clean starting level; the dotted line marks the $0.99$
ceiling most clean rows already sit on.}
\label{fig:signatures}
\end{figure*}

To keep the comparison interpretable, the inconsistent returns must remain schema-valid. Contamination has to be invisible to any channel that does not consult the trajectory, since a corruption already improbable under the schema alone carries the signature of an execution failure and the two families stop being separated by the manipulation. Schema validation intercepts out-of-schema values, which is why \Astat{} is restricted to the enumeration of legal statuses, but validation alone is not sufficient: $\Sz$ is a likelihood over the serialized payload rather than a validator, and it will register a record whose fields are individually legal yet jointly incoherent. We therefore report $\Sz$ alongside the other quantities instead of thresholding it, so that whether the manipulation stayed inside the schema channel is visible in Fig.~\ref{fig:signatures} rather than asserted here. Only the returned payload is perturbed, so the database and every other tool continue to report true values and the contradiction is genuine.

Tasks are included in the paired set if their reference action sequence is non-empty, contains a valid injection site with at least one later action, and can be replayed without error. We also exclude tasks whose clean baseline already conflicts with the user's stated beliefs. In total, 53 of the 114 tasks satisfy these conditions. The two error arms are included independently when their target fields are available. Therefore, the 189 scored decision points are not simply $53 \times 4$. Table~\ref{tab:arms} reports the number of cases in each arm.

We first measure each quantity at the decision point immediately after the injection. Since the injected return remains in the context, we also repeat the measurement along the later trajectory. Let $s$ denote the injected call and let $\koff$ count calls relative to it, with $\koff=0$ denoting the injected call. At step $\koff$, $\Ggap$ is computed for the return of call $s+\koff$, while $\pi$ is measured at the next decision point that issues call $s+\koff+1$. This is the first decision point at which that return appears in the context. Only the payload at $\koff=0$ is modified, and the database remains unchanged. Thus, all returns for $\koff\geq1$ are byte-identical across the four arms. All prefixes for $\koff\leq-1$ are also identical, so the four curves coincide before the injection.

All quantities are scored with Qwen3-8B~\cite{qwen3} in \texttt{bfloat16} on NVIDIA A40, using HuggingFace transformers 5.15.0 with
PyTorch 2.6.0+cu124, and scoring runs at batch size one with
\texttt{sdpa} attention, TF32 disabled, and KV caching disabled.

\section{Experimental Results}
\label{sec:results}

\begin{figure*}[!t]
\centering
\includegraphics[width=\textwidth]{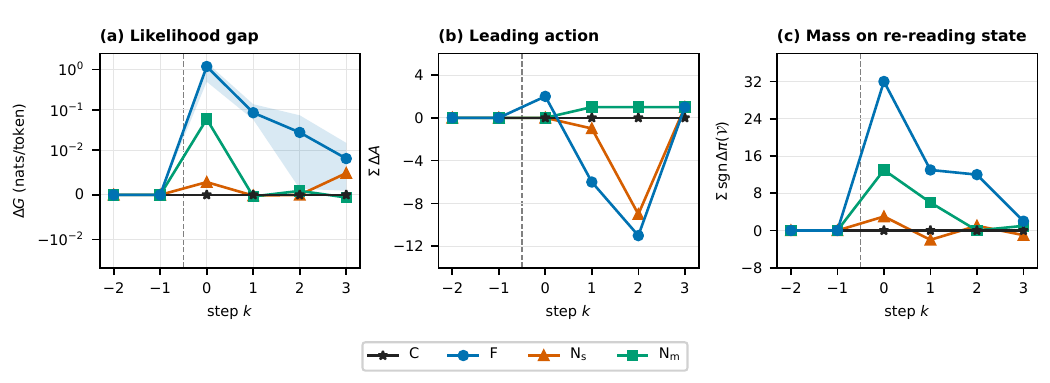}
\caption{The same readings along the trajectory. Every quantity is paired against the same task's \Acl{} row at the same step; $\koff$ counts calls from
the injected one, so returns at $\koff\geq1$ are byte-identical across arms and
the arms coincide left of the dashed line. \textbf{(a)} $\Delta\Ggap$, medians on
a symmetric-log axis, with the interquartile range of \Afail{} shaded.
\textbf{(b)} Net paired change in whether the leading action is the reference
action; negative is away from it. \textbf{(c)} Net paired direction of $\Pver$,
the mass on the two tools that re-read state, as tasks up minus tasks down.
Counts in Table~\ref{tab:offset}.}
\label{fig:offset}
\end{figure*}

We evaluate whether the proposed signals can distinguish different types of tool failures and capture their effects on subsequent agent decisions. We first examine the likelihood-based signals at the injection point and along the following trajectory. We then study how each failure changes the agent's action distribution. Finally, we compare the two groups of signals to determine what information each captures.


\subsection{The likelihood channel}
\label{sec:obs-gap}

We first examine whether the likelihood-based signals distinguish the three failure conditions. $\Ggap$ is negative for all 53 clean returns, showing that the additional trajectory context consistently makes clean returns easier to predict. We therefore use the paired change from the clean return to measure the effect of each injected failure.


\paragraph{Schema-level differences.}
Fig.~\ref{fig:signatures}(a) compares each injected arm with its clean counterpart, which is placed at the origin. \Afail{} shows a clear shift along the schema axis. $\Delta\Sz$ is positive for all 53 rows and reaches a region that no other arm enters. In contrast, the two inconsistent arms remain close to zero and overlap with the variation already present among clean rows. This confirms that the failed return is abnormal under the tool schema, while the schema-valid inconsistent returns are not. The separation is not caused by return length. Although \Afail{} is a short error string and the other arms contain full records, $\Sz$ is computed as a per-token mean. Scoring only the first token of each return gives the same separation.

\paragraph{Context-level differences.}
The gap axis shows a different pattern. \Afail{} increases $\Ggap$ on 45 of 53 rows, as expected for a return that is unlikely under both conditionings. Among the inconsistent arms, \Apay{} increases $\Ggap$ on 30 of 37 rows, while \Astat{} shows no clear direction. This difference depends on whether the conflicting value is already present in the prefix. For \Apay{}, $\Ggap$ increases on 30 of the 31 rows where the correct identifier appears verbatim earlier in the trajectory. It does not increase on any of the six rows where \texttt{get\_user\_details} has not yet been called. In contrast, the contradiction in \Astat{} often has to be derived from the domain policy. Among the 14 policy-dependent cases, 6 remain flat, similar to the 8 rows where the prefix does not constrain the value. These results suggest that $\Ggap$ is sensitive to direct conflicts with values already written in the context, but is less effective when the conflict requires policy reasoning.

\paragraph{Effects along the trajectory.}
We test whether these likelihood effects remain visible after the injected return has entered the context. For $\koff\geq1$, the scored return is byte-identical across all arms. Its likelihood under the tool schema is therefore also identical. In this setting, $\Delta\Ggap$ reduces to $\Delta\So$, so the remaining difference comes only from the changed trajectory prefix.

The two failure families show different patterns under this control. At $\koff=1$, \Afail{} increases $\Ggap$ on all 51 available tasks, with a median change of $+0.085$ nats per token and an interquartile range of $[+0.058,+0.138]$. A sign test gives $p<10^{-15}$. In contrast, neither inconsistent arm shows a consistent shift. \Astat{} is positive on 22 of 45 rows and \Apay{} on 18 of 37 rows, with median changes of $-0.0002$ and $-0.0004$. The effect of \Afail{} then decreases over later calls, with median changes of $+0.028$ at $\koff=2$ and $+0.008$ at $\koff=3$, while the inconsistent arms remain close to zero. The same pattern holds when we restrict the analysis to the 19 tasks that reach every step.

These results show that the likelihood channel captures two different effects. At the injection site, a schema-valid falsified field is visible only when its correct value already appears explicitly in the prefix. After the injection, only the failed return leaves a consistent likelihood effect on the later trajectory.

\begin{table}[t]
\centering
\caption{Paired movement per arm at the injection site, counted against each
task's own clean row \Acl. Entries are row counts; the right-hand block restricts
to the tasks whose clean entropy clears $0.01$ nats, and $H$ counts those falling
back below it.}
\label{tab:arms}
\setlength{\tabcolsep}{4pt}
\begin{tabular}{@{}lcccccc@{}}
\toprule
& \multicolumn{3}{c}{All rows} & \multicolumn{3}{c}{Entropy headroom} \\
\cmidrule(lr){2-4}\cmidrule(lr){5-7}
Arm & $n$ & $\Delta\Sz>0$ & $\Delta\Ggap>0$ & $n$ & $H<0.01$ & action changes \\
\midrule
\Afail & 53 & 53 & 45 & 22 & 15 & 15 \\
\Astat & 46 & 25 & 26 & 20 &  8 &  6 \\
\Apay  & 37 & 18 & 30 & 16 &  2 &  0 \\
\bottomrule
\end{tabular}
\end{table}

\subsection{The action channel}
\label{sec:obs-action}

We next examine how the injected failures affect the agent's action distribution. We consider three properties of $\pi$, namely its entropy, its most probable action, and the probability mass assigned to different action groups.

\paragraph{Action uncertainty.}
The clean action distributions are already highly concentrated at the injection site. Their median entropy is about $0.005$ nats, compared with the maximum value of $\log 17 \approx 2.83$. Among the 53 clean rows, 31 have entropy below $0.01$ nats, where $\pi$ is already close to a point mass. We therefore focus on the remaining 22 rows with enough room for a meaningful shift. Fig.~\ref{fig:signatures}(b) shows that \Astat{} produces the clearest increase in uncertainty. Two \Astat{} rows move from below $0.01$ nats to above $0.5$ nats, including one with $\Delta H = +1.20$. In comparison, \Afail{} reaches at most $+0.22$ from the same range, while \Apay{} reaches only $+0.002$. After the injection site, however, the entropy effect quickly disappears. The median $\Delta H$ remains within $0.0011$ nats of zero for all arms and all later steps. Action entropy therefore captures only a limited part of the immediate response to the injected failure.

\paragraph{Most probable action.}
We next examine whether the most probable action changes relative to the reference action. Among the 22 rows with sufficient entropy headroom, the leading action changes in 15 cases under \Afail{}, 6 under \Astat{}, and none of the 16 available cases under \Apay{}. A similar pattern appears later in the trajectory. At $\koff=1$, \Afail{} moves $\amax$ away from the reference action on 9 tasks and toward it on 3. At $\koff=2$, the corresponding counts are 14 and 3. \Astat{} shows a weaker and delayed effect, with 10 moves away from the reference action and 1 move toward it at $\koff=2$. \Apay{} does not change the leading action at any step. Since the clean agreement rate varies across steps, we report only paired changes. These results show that \Afail{} and \Astat{} can both alter the leading action, while \Apay{} has little effect. However, the leading action alone does not reveal where the probability mass moves.

\paragraph{Action-group probability.}
We therefore examine the probability mass assigned to different action groups. On clean returns, $\Pver$ is strongly bimodal. Among the 53 rows, 35 have $\Pver>0.99$, while 14 have $\Pver<0.01$. Fig.~\ref{fig:signatures}(c) therefore compares each injected arm with its own clean starting point. On the 18 tasks where the clean distribution leaves room to move, \Afail{} increases $\Pver$ in every case, often from below one half to almost one. Across all 53 rows, $\Pver$ increases on 34 and decreases on only 2. In contrast, \Astat{} shows no clear direction, increasing $\Pver$ on 17 rows and decreasing it on 14. $\Pact$ is below $0.01$ on 46 of the 53 clean rows, and $\Presp$ is smaller still, so neither provides a reliable signal at the injection site.
The same difference continues after the injection. For \Afail{}, the net increase in $\Pver$ is $+13$ tasks at $\koff=1$ and $+12$ at $\koff=2$, before decreasing to $+2$ at $\koff=3$. \Astat{} shows no consistent change at later steps, as shown in Fig.~\ref{fig:offset}(c).

The destination of the changed action further separates the two conditions. Across all decision points from the injection onward, \Afail{} moves $\amax$ to a tool that re-reads state in 16 of 28 cases. Fourteen of these are repeated reads of the same order whose earlier call failed. In contrast, none of the 15 changes under \Astat{} moves to a verification tool. Instead, 9 move to a write action, including 8 moves to the same write tool. This result shows that \Afail{} tends to push the agent toward verification, while \Astat{} can redirect the agent toward state-changing actions.

\begin{table}[t]
\centering
\caption{Paired movement at each step from the injection site. The first block
counts rows whose gap rose, out of the rows where the step exists; those
denominators apply to the two blocks below, which give net task counts.}
\label{tab:offset}
\setlength{\tabcolsep}{5pt}
\begin{tabular}{@{}llcccc@{}}
\toprule
 & Arm & $\koff=0$ & $\koff=1$ & $\koff=2$ & $\koff=3$ \\
\midrule
\multirow{3}{*}{$\Delta\Ggap>0$}
       & \Afail & 45/53 & 51/51 & 32/41 & 14/19 \\
       & \Astat & 26/46 & 22/45 & 19/39 & 12/19 \\
       & \Apay  & 30/37 & 18/37 & 22/33 & 6/17 \\
\midrule
\multirow{3}{*}{$\Sigma\Delta A$}
       & \Afail & $+2$ & $-6$ & $-11$ & $+1$ \\
       & \Astat & $0$  & $-1$ & $-9$  & $+1$ \\
       & \Apay  & $0$  & $+1$ & $+1$  & $+1$ \\
\midrule
\multirow{3}{*}{$\Sigma\,\mathrm{sgn}\,\Delta\Pver$}
       & \Afail & $+32$ & $+13$ & $+12$ & $+2$ \\
       & \Astat & $+3$  & $-2$  & $+1$  & $-1$ \\
       & \Apay  & $+13$ & $+6$  & $0$   & $+1$ \\
\bottomrule
\end{tabular}
\end{table}

\subsection{Comparison of Failure Signals}
\label{sec:obs-cross}

The likelihood-based and action-based signals capture different aspects of the three failure conditions. \Afail{} is visible in both groups of signals. It is the only condition that clearly departs from the schema-level baseline, remains visible in the likelihood comparison after the injection, and consistently shifts probability toward tools that re-read state.

The two inconsistent conditions show different patterns. \Apay{} is detected mainly by the likelihood comparison at the injection site because the correct value already appears explicitly in the prefix. However, it produces little change in the following action distribution. \Astat{} shows the opposite behavior. It is not clearly detected by the likelihood comparison either at the injection site or later in the trajectory, but it can still increase action uncertainty and change the leading action. At $\koff=2$, its effect on the leading action is close to that of \Afail{}.

The destination of these changes is also different. When \Afail{} changes the leading action, it usually moves the agent back toward a tool that re-reads state. In contrast, \Astat{} often moves the agent toward a state-changing action, with eight of nine such changes going to the same write tool. These results show that a failure can have a strong effect on the agent's behavior even when it is weak under the likelihood-based signals. The two groups of signals are therefore complementary, and no single quantity captures all three failure conditions.

\section{Discussion}
\label{sec:discussion}

Each channel is a function of something different, and its blind spot follows
from that. $\Ggap$ compares two likelihoods over the same string, so it can
register a value the prefix has already written out but not one the prefix rules
out through the domain policy; the action distribution is a function of the state
the return implies rather than of the return itself, and so orders the arms by
how much of the option set changes, indifferent to why the return is wrong.
Neither, therefore, is a function of the observation and the policy jointly. A payload that contradicts nothing yet still misleads, such as a plausible note about a return policy no other tool can confirm, should on this account leave no trace in either channel~\cite{injecagent}, and separating the merely contradictory from the consequential needs a quantity of that joint kind. The two families not injected here follow from the same account: an irrelevant return carries content the trajectory does not anticipate, so the likelihood comparison should move on it much as it moves on an inconsistent one, while an ambiguous one contradicts nothing and should leave that comparison where it found it, being the one family with a reason to disperse the action distribution rather than concentrate it. Both are predictions, not results.

The curves of Fig.~\ref{fig:offset} capture the prefix-mediated component of the degradation an agent may show after receiving a faulty return, rather than the full degradation of a live rollout. In a live rollout, that degradation has two sources: the altered context itself and the agent's subsequent actions in response to it. Replaying the reference sequence fixes the action at every step and removes the second source. This makes the trace at $\koff\geq1$ attributable to the changed prefix, but also prevents it from being interpreted as a task-completion effect. How far the forced
trajectory has drifted from what the model would have done is measured rather
than assumed: the reference action stops being the most probable one on a third
to a half of rows by $\koff=2$ even on the clean arm, which is where we stop
reading the tail.

The analysis has several limitations. We score decision points under teacher forcing, so the results reflect the agent's immediate response rather than its final task outcome. Each failure arm is included only when its target field is available, so the counts are exact within each arm but only indicative across arms. The experiments are also limited to one domain, one model, one injection site, and template-based rather than adversarial noise.
\section{Conclusion}
We propose a tool-return boundary analysis that combines likelihood-based and action-based signals to distinguish incomplete and inconsistent tool failures. We score each return under both the tool schema and the full trajectory prefix, and we measure how the return changes the agent's distribution over legal actions. Experiments on the retail domain of $\tau^2$-bench show that the likelihood comparison clearly captures incomplete returns and inconsistencies that directly conflict with values already present in the context. In contrast, action-based signals capture how strongly a faulty return changes the next decision, including cases that the likelihood comparison misses. In particular, a falsified status can redirect the agent toward a state-changing action even when it produces little likelihood signal, while a failed call more often shifts the agent back toward state verification. These results show that the two signal groups provide complementary information at the tool-return boundary, and that reliable failure diagnosis requires considering both the form of the returned content and its effect on subsequent actions.



\begin{thebibliography}{00}

\bibitem{agentnoisebench}
R. Wang, Y. Chen, Y. Wang, C. Wu, J. Fang, X. Cai, Q. Gu, H. Su, A. Zhang, X. Wang, X. Cai, and T.-S. Chua, ``AgentNoiseBench: Benchmarking robustness of tool-using LLM agents under noisy condition,'' in \emph{Proc. 43rd Int. Conf. Mach. Learn. (ICML)}, Seoul, South Korea, 2026.

\bibitem{actundernoise}
Y. Chen, X. Cai, J. Fang, Z. Han, Y. Wang, Y. Shi, Y. Zhang, Q. Gu, X. Cai, X. Wang, A. Zhang, and T.-S. Chua, ``Learning to act
under noise: Enhancing agent robustness via noisy environments,'' arXiv preprint arXiv:2605.27209, 2026.

\bibitem{L2}
T. Li, R. Huang, L. Chen, C. S. Jensen, and T. B. Pedersen, ``Compression of uncertain trajectories in road networks,'' \emph{Proc. VLDB Endow.}, vol. 13, no. 7, pp. 1050--1063, Mar. 2020.

\bibitem{L3}
T. Li, L. Chen, C. S. Jensen, and T. B. Pedersen, ``TRACE: Real-time compression of streaming trajectories in road networks,'' \emph{Proc. VLDB Endow.}, vol. 14, no. 7, pp. 1175--1187, Mar. 2021.

\bibitem{toolhalluc}
H. Xu, Z. Zhu, L. Pan, Z. Wang, S. Zhu, D. Ma, R. Cao, L. Chen, and K. Yu, ``Reducing tool hallucination via reliability alignment,'' in \emph{Proc. 42nd Int. Conf. Mach. Learn. (ICML)}, Vancouver, Canada, pp. 69992--70006, 2025.

\bibitem{toolbehonest}
Y. Zhang, J. Chen, J. Wang, Y. Liu, C. Yang, C. Shi, X. Zhu, Z. Lin, H. Wan, Y. Yang, T. Sakai, T. Feng, and H. Yamana, ``ToolBeHonest: A multi-level hallucination diagnostic benchmark for tool-augmented large language models,'' in \emph{Proc. 2024 Conf. Empirical Methods Nat. Lang. Process. (EMNLP)}, Miami, Florida, USA, pp. 11388--11422, 2024.

\bibitem{L5}
Y. Yao, L. Chen, Z. Fang, Y. Gao, C. S. Jensen, and T. Li, ``Camel: Efficient compression of floating-point time series,'' \emph{Proc. ACM Manag. Data}, vol. 2, no. 6, Art. no. 227, pp. 1--26, Dec. 2024.

\bibitem{L7}
Y. Yao, H. Jie, L. Chen, T. Li, Y. Gao, and S. Wen, ``TSec: An efficient and effective framework for time series classification,'' in \emph{Proc. 40th IEEE Int. Conf. Data Eng. (ICDE)}, Utrecht, Netherlands, pp. 1394--1406, 2024.

\bibitem{predentropy}
A. Malinin and M. Gales, ``Uncertainty estimation in autoregressive structured prediction,'' in \emph{Proc. 9th Int. Conf. Learn. Represent. (ICLR)}, Virtual Event, Austria, 2021.

\bibitem{semanticentropy}
S. Farquhar, J. Kossen, L. Kuhn, and Y. Gal, ``Detecting hallucinations in large language models using semantic entropy,'' \emph{Nature}, vol. 630, no. 8017, pp. 625--630, Jun. 2024.

\bibitem{react}
S. Yao, J. Zhao, D. Yu, N. Du, I. Shafran, K. Narasimhan, and Y. Cao, ``ReAct: Synergizing reasoning and acting in language models,'' in \emph{Proc. 11th Int. Conf. Learn. Represent. (ICLR)}, Kigali, Rwanda, 2023.

\bibitem{L6}
D. Hu, Z. Fang, H. Fang, T. Li, C. Shen, L. Chen, and Y. Gao, ``Estimator: An effective and scalable framework for transportation mode classification over trajectories,'' \emph{IEEE Trans. Intell. Transp. Syst.}, vol. 25, no. 11, pp. 15562--15573, Nov. 2024.

\bibitem{tau2bench}
V. Barres, H. Dong, S. Ray, X. Si, and K. Narasimhan,
``$\tau^2$-Bench: Evaluating conversational agents in a dual-control environment,'' in \emph{Proc. 43rd Int. Conf. Mach. Learn. (ICML)}, Seoul, South Korea, 2026.

\bibitem{taubench}
S. Yao, N. Shinn, P. Razavi, and K. Narasimhan, ``$\tau$-bench: A benchmark for tool-agent-user interaction in real-world domains,'' in \emph{Proc. 13th Int. Conf. Learn. Represent. (ICLR)}, Singapore, pp. 74824--74876, 2025.

\bibitem{agenticrl}
G. Zhang, H. Geng, X. Yu, Z. Yin, Z. Zhang, Z. Tan, H. Zhou, Z.-Z. Li, X. Xue, Y. Li, Y. Zhou, Y. Chen, C. Zhang, Y. Fan, Z. Wang, S. Huang, F. P. Velez, Y. Liao, H. Wang, M. Yang, H. Ji, J. Wang, S. Yan, P. Torr, and L. Bai, ``The landscape of
agentic reinforcement learning for LLMs: A survey,'' \emph{Trans. Mach. Learn. Res.}, Jan. 2026.

\bibitem{qwen3}
A. Yang, A. Li, B. Yang, B. Zhang, B. Hui, B. Zheng, B. Yu, C. Gao, C. Huang, C. Lv, C. Zheng, D. Liu, F. Zhou, F. Huang, F. Hu, H. Ge, H. Wei, H. Lin, J. Tang, J. Yang, J. Tu, J. Zhang, J. Yang, J. Yang, J. Zhou, J. Zhou, J. Lin, K. Dang, K. Bao, K. Yang, L. Yu, L. Deng, M. Li, M. Xue, M. Li, P. Zhang, P. Wang, Q. Zhu, R. Men, R. Gao, S. Liu, S. Luo, T. Li, T. Tang, W. Yin, X. Ren, X. Wang, X. Zhang, X. Ren, Y. Fan, Y. Su, Y. Zhang, Y. Zhang, Y. Wan, Y. Liu, Z. Wang, Z. Cui, Z. Zhang, Z. Zhou, and Z. Qiu, ``Qwen3 technical report,'' arXiv preprint arXiv:2505.09388, 2025.

\bibitem{injecagent}
Q. Zhan, Z. Liang, Z. Ying, and D. Kang, ``InjecAgent: Benchmarking indirect prompt injections in tool-integrated large language model agents,'' in \emph{Findings Assoc. Comput. Linguistics: ACL 2024}, Bangkok, Thailand, pp. 10471--10506, 2024.


\end{thebibliography}
\end{document}